\pdfoutput=1
\documentclass[11pt,dvipsnames]{article}

\usepackage{acl}
\usepackage{times} 
\usepackage{latexsym}
\usepackage[T1]{fontenc}
\usepackage[utf8]{inputenc}
\usepackage{microtype}
\usepackage{graphicx} 
\usepackage{comment}
\usepackage{expex} 
\lingset{
aboveglftskip= -0.2cm,
aboveexskip = 0.2cm,
belowexskip = 0.2cm,
interpartskip = 0.1cm,
}

\usepackage{caption} 
\usepackage{graphicx, subcaption, float}
\usepackage{color, colortbl}
\definecolor{LightBlue}{rgb}{0.75, 1, 0.9}
\definecolor{LightGreen}{rgb}{0.8, 0.8, 0.5}
\usepackage{adjustbox}

\usepackage{settings_xtong}

\title{A framework for annotating and modelling intentions behind metaphor use}

 \author{Gianluca Michelli\textsuperscript{*} \And Xiaoyu Tong\textsuperscript{*} \\
 ILLC, University of Amsterdam, the Netherlands \\
         \texttt{gianlucamichelli@gmail.com} \\ \texttt{x.tong@uva.nl} \\ \texttt{e.shutova@uva.nl} \And Ekaterina Shutova }

\begin{document}

\maketitle

\newcommand\blfootnote[1]{
    \begingroup
    \renewcommand\thefootnote{}\footnote{#1}
    \addtocounter{footnote}{-1}
    \endgroup
}

\blfootnote{\textsuperscript{*}These authors contributed equally to this work.}

\begin{abstract}

Metaphors are part of everyday language and shape the way in which we conceptualize the world. Moreover, they play a multifaceted role in communication, making their understanding and generation a challenging task for language models (LMs). While there has been extensive work in the literature linking metaphor to the fulfilment of individual intentions, no comprehensive taxonomy of such intentions, suitable for natural language processing (NLP) applications, is available to present day. In this paper, we propose a novel taxonomy of intentions commonly attributed to metaphor, which comprises 9 categories. We also release the first dataset annotated for intentions behind metaphor use.
Finally, we use this dataset to test the capability of large language models (LLMs) in inferring the intentions behind metaphor use, in zero- and in-context few-shot settings.
Our experiments show that this is still a challenge for LLMs.
 
\end{abstract}

\section{Introduction}

Metaphors are pervasive in literary and political discourse, but they are also frequently used in our everyday language. Therefore, they need to be interpreted by natural language understanding systems.
Consider the following quote from the “I Have A Dream” speech by Dr. Martin Luther King, Jr.: \textit{Now is the time to \underline{rise} from the \underline{dark} and \underline{desolate valley} of segregation to the \underline{sunlit path} of racial justice}. In this sentence, several words are used metaphorically, among which are \textit{dark} and \textit{sunlit}. According to the Conceptual Metaphor Theory (CMT), a single \textit{conceptual metaphor} may underpin these diverse linguistic manifestations \cite{LJ80}.
Conceptual metaphors are mappings that
allow one to conceptualize a TARGET domain (often more complex or abstract) based on prior knowledge of a SOURCE domain (more concrete). For instance, the conceptual metaphor JUSTICE IS LIGHT allows one to understand the abstract domain of racial justice (the TARGET) in terms of the more concrete domain of light (the SOURCE). Thus, segregation is associated with a dark, gloomy place, while social justice is a bright, sunny one. On a higher level of analysis, these metaphors are used in Dr. King's speech with specific communicative goals: making complex issues intelligible, appealing to the audience's emotions and calling them to action, and possibly more.

Following the rise of CMT, metaphor theorists have increasingly focused their research on the many and varied effects that metaphor has on cognitive processes. 
It has been observed that in some recurring contexts, \textit{e.g.} in political speech, metaphorical language tends to be preferred to literal language due to its effects on the receivers  \cite{Musolff2004}. This has led some researchers, most notably \citet{Steen2008, Steen2023}, to emphasise the \textit{communicative dimension} of metaphor, a dimension in which metaphors are sometimes used deliberately to produce specific effects. Computational linguists have also investigated pragmatic aspects of metaphor, such as its affective component \cite{SW2022} and argumentative potential \cite{KF2013}.
The communicative role of metaphors can be  explained in terms of the intentions (\textit{viz}. discourse goals) that they are supposed to achieve. The literature relating metaphor and intention is rich but generally quite fragmented. With some exceptions \cite{RK1994}, metaphor scholars tend to focus only on isolated intentions. Hence, there is still a lack of a systematic and comprehensive account of intentions behind metaphor use and an operationalised framework enabling annotation of such intentions in linguistic data. 

In this paper, we fill in this gap by systematising the existing literature on metaphor and intention, and proposing a first-of-a-kind unified taxonomy of intentions behind metaphor use. We further propose an annotation procedure and release a first dataset annotated for intentions behind metaphor use. We show that the proposed taxonomy is thus suitable for annotating metaphors in unrestricted text. 
The dataset is publicly available on \url{https://github.com/GMichelli/intentions-behind-metaphor}.

Using this dataset, we test GPT-4 Turbo and two Llama2-Chat models (the 13B and 70B versions)
on their ability to infer the intentions behind metaphor use.
The task requires the models to select one category from the taxonomy for a given metaphorical expression in a sentence.
The best-performing model, GPT-4, reaches an average accuracy of 42.99\% in the zero-shot setting and a slightly higher accuracy of 44.68\% in the five-shot setting, demonstrating that inferring the intentions behind metaphor use is a challenging task for state-of-the-art LLMs.

\section{Related work}

\subsection{Conceptual and deliberate metaphor}

In a seminal paper, Ortony emphasized the necessary role of metaphor in everyday language \cite{Ortony1975}. Proponents of CMT reinforced this idea, highlighting that our own conceptual system is, at least partly, metaphorically structured \cite{LJ80}. Abstract concepts, \textit{e.g.} emotions like love, are conceived of through various kinds of conceptual metaphors.

While CMT revealed the pervasive nature of metaphor in human cognition, several authors emphasized the importance of communicative aspects in the analysis of metaphors. In particular, \citet{Steen2008} stressed the significance of discerning deliberate metaphors from non-deliberate ones. Deliberate Metaphor Theory (DMT) departs from CMT by recognizing that only metaphors intentionally used \textit{as} metaphors in communication involve online cross-domain mappings \cite{Steen2017, Steen2023}, and non-deliberate metaphors can be processed differently--by lexical disambiguation. DMT faced criticisms, however. For instance, \citet{Gibbs2011} highlighted the difficulty of identifying deliberate metaphors without specific linguistic markers and the unreliability of producers' conscious judgments on their own intentions. In order to address these challenges, advocates of DMT developed the Deliberate Metaphor Identification Procedure (DMIP) and clarified the distinction between deliberate and conscious use of metaphors \cite{Reijnierse2018, Steen2014}.

\subsection{Intentions in language use}

The notion of communicative intention (CI) holds a central position in the field of pragmatics. CI is the speaker's intention to convey non-natural meaning through their utterances \cite{Grice1957}. In its original formulation, it comprises three sub-intentions \cite{Recanati1986}: (i) the intention to produce a specific effect in the hearer, (ii) the intention that the hearer recognizes (i), and (iii) the intention for (i) to be fulfilled, at least in part, by its recognition. Subsequent research has shown that one can distinguish among different intentions, varying in nature--prior intention \textit{vs} intention in action \cite{Searle1983}, temporal aspect--proximal \textit{vs} prospective \cite{HJ2012}, and social dimension--individual \textit{vs} social \cite{Ciaramidaro2007}. 

As stressed by \citet{Gibbs1999}, conceiving of intentions as individual mental states makes them opaque since agents are not always aware of the causes of their behaviour. Intentions should be viewed as a \textit{social judgment} instead. Inspired by Anscombe's philosophy of action \cite{Anscombe1957}, in this paper we follow a “grammatical” approach and conceive of intentions as features attributed to linguistic acts. More specifically, intentions are those \textit{reasons} that speakers may provide once asked why they resorted to certain metaphors. Ultimately, they serve as hermeneutic tools for understanding human behaviour broadly, and linguistic behaviour in particular.

\subsection{Intentions and metaphor}

Although there is not a common notion of intention shared among all metaphor scholars, in the literature intentions are typically formalized as prior intentions, that is, as representations in the speaker’s mind of their goals. A paper by Roberts and Kreuz builds a first taxonomy of intentions for various forms of figurative language, including metaphor \cite{RK1994}. This taxonomy was developed through experiments where participants were asked to provide reasons for using each figure of speech. We believe that the taxonomy has some limitations. First, the participants’ judgments may have been biased by a comparativist definition of metaphor (\textit{i.e.} metaphor as implicit comparison). Second, the number of participants assigned to metaphor was too low. As a result, there is still the need for an improved, metaphor-specific taxonomy.

Previous work has explored to some extent the relation between metaphor and individual intentions. Researchers have observed how metaphors can convey emotions \cite{FO1987, fussell-moss-2014-figurative}, persuade \cite{SD2002, VanStee2018}, contribute to argumentation \cite{Wagemans2016, VanPoppel2021}, serve didactic purposes \cite{Cameron2003}, add humor \cite{attardo-2015-humorous}, and cultivate intimacy among speakers and comprehenders \cite{Cohen1978, Goatly1997}. These studies highlight the multifaceted nature of intentions behind the use of metaphor in communication and inspired our novel taxonomy.

\section{Taxonomy of intentions}

We introduce the individual intention categories, motivating them through theoretical considerations, previous literature and examples adapted from published material.

\paragraph{Lexicalized metaphor.} These metaphors are associated with a plain communicative intention, and the utterance is judged as meant to convey just its propositional message. For lexicalized metaphors, the question of why a metaphor was preferred over a literal paraphrase does not arise in interpretation. In Cameron’s words, the metaphoric expression is ”just the way to say it” \cite{Cameron2003}.

\pex \label{ex: lexicalized_met}
\a I \underline{fell in} love.
\a Summer \underline{bedding} is looking tired.
\xe 

Sentence (\ref{ex: lexicalized_met}a) is an example of how the language of emotions often relies on metaphors. This observation, already noted by \citet{FO1987}, aligns with the idea that emotions may be conceptualized metaphorically, as maintained by CMT. Example (\ref{ex: lexicalized_met}b), instead, shows how the language we use to talk about some activities tends to have its own metaphorical jargon. This is true for academic domains such as mathematics, physics and the like, but also for non-academic domains like sports or hobbies. Both examples are cases of lexicalized metaphors which constitute the most conventional way of talking about the TARGET.

\paragraph{Artistic use of metaphor.} These metaphors are used to attribute at once a whole set of features to the TARGET. These features need not be  clearly determined in advance. Ultimately, the intention is to stimulate the receiver’s creative interpretation.

\pex \label{ex: Artistic_met}
\a It is the east, and Juliet is \underline{the Sun}.
\a Fermi’s \underline{mantle} in physics had fallen on his young shoulders.
\xe 

Some metaphors are not easily paraphrasable because they could be paraphrased in a number of different, yet equally valid, ways. The ambiguity of the metaphorical meaning can be inherent to the TARGET of the metaphor or it can be related to the set of features that the metaphor attributes. At least in  poetry and literature, interpreters tend to activate multiple mappings at once \cite{Rasse2020} and ambiguity in interpretation is shown to correlate with aesthetic liking \cite{JK2017}.

\paragraph{Visualization.} The utterer might resort to a metaphor whose SOURCE is easier to visualize than the TARGET. The intention is to help the receiver to form an intuitive representation of the latter.

\pex \label{ex: Visualization}
\a It was \underline{like a very bright light was just} \underline{shining outward}.
\a It would bounce up and down \underline{like a} \underline{yo-yo}.
\xe 

Metaphors often hinge on a highly concrete/imaginable SOURCE to address an abstract TOPIC\footnote{In psycholinguistics literature, imageability refers to the property of words to easily evoke a mental image of their meaning \cite{Paivio1968}. Imageability and concreteness, thought positively correlated, might be two distinct constructs \cite{Dellantonio2014, gargett-barnden-2015-modeling}.}. This is particularly true for subjective feelings, as in example (\ref{ex: Visualization}a). \citet{fussell-moss-2014-figurative} provide evidence for the ability of metaphors to express precise emotional states. More recently, \citet{Broadwell2013} developed a prototype model for automated metaphor identification partly based on imageability.

Some metaphors do not constitute mappings from the concrete to the abstract, but just from the familiar to the unfamiliar (\ref{ex: Visualization}b). As already stressed by \citet{Ortony1975}, metaphoric expressions are often perceived as more vivid than their literal paraphrases. Thus, they can foster the formation of a more insightful mental image. Vivid metaphors can be instrumental not only for descriptive purposes. As reported in \cite{Cameron2003}, they can also be used to express more clearly some commands (\textit{cf.} a PE teacher explaining their pupils how to perform a dance: \textit{you are \underline{spokes in a wheel}}).

\paragraph{Persuasiveness.} Using a metaphor to refer to the TARGET---in a political speech, for instance---the author can give it a non-neutral connotation. This connotation is not motivated by explicit arguments. The intention is for the audience to adopt the utterer’s perspective or stance towards the TARGET.

\pex \label{ex: persuasiveness}
\a The islamic \underline{wave}.
\a This \underline{slender} and \underline{anaemic} first novel by a notable poet.
\xe 

As already stressed by \citet{LJ80}, metaphors generally highlight some aspects of the TARGET, while at the same time hiding others. This process of highlighting and hiding causes a \textit{framing effect} on the receiver, whereby the TARGET is seen, as it were, through the distorting lens of the SOURCE. The availability of several experiments and of meta-studies \cite{SD2002, VanStee2018} makes the Persuasiveness category one that is most supported empirically. 

\paragraph{Explanation.} This type of metaphors are used for didactic purposes. The intention is to explain a new or already familiar concept to the addressee. There is some knowledge asymmetry in the discourse from specialists to non-specialists, \textit{e.g.} from teacher to students.

\pex \label{ex: Explanation}
\a The atmosphere is \underline{the blanket} of gases that surrounds the earth.
\a When the neutron falls apart, \underline{spits out} an electron, it becomes a proton.
\xe 

The clarifying effect of metaphor has been recognized in the existing study of intentions behind it by \citet{RK1994}. The role metaphors play in educational settings--\textit{viz.} in primary education--has been analysed in detail by \citet{Cameron2003}. Moreover, there is some empirical evidence for the usefulness of certain (deliberate) metaphors in undergraduate lectures \cite{BJ2015}. However, the use of metaphors in education does not go without risks of blocking further understanding, as highlighted by \citet{Spiro89}.

\paragraph{Argumentative metaphor.} These metaphors are part of explicit arguments intended by the author to convince the audience of a certain claim. The intention is to make the argument more compelling.

\pex \label{ex: Argumentative_met}
\a But the villages are dying, becoming suburbs
or \underline{dormitories} where few people
work but many sleep.
\a If so, it will be a gamble, because he \underline{flopped on} his only previous international appearance in Saudi Arabia.
\xe 

As pointed out, among others, by \citet{VanPoppel2021}, argumentative metaphors can be used to make an effective statement, either as a standpoint or as a starting point (premise) for an argument. Moreover, they can also actively contribute to the flow of argumentation (\ref{ex: Argumentative_met}a,b).

\paragraph{Social interaction.} These metaphors focus on interpersonal relations, group or cultural conventions and the like. The intention is to create or strengthen some bond between producer and receiver.

\pex \label{ex: Social_interaction}
\a \underline{Sleepy} Joe, \underline{Crooked} Hillary.
\a She \underline{passed away}.
\xe 

A metaphor can bring closer its maker and appreciators in a number of different ways. First, it can exploit the fact that they belong to the same group--\textit{e.g.}, Trump's supporters (\ref{ex: Social_interaction}a). In such cases, a social metaphor is used to isolate the desired receiver from the general public \cite{Cohen1978}, thus reinforcing the in-group/out-group dynamic. Second, metaphor can be used to conceal a TARGET that is experienced as negative. If they understand this, the receiver becomes aware of the additional care put by the producer in their utterance. The shared awareness fosters intimacy building between the pair and stimulates empathetic effects (\ref{ex: Social_interaction}b).

\paragraph{Humour.} The intention is to entertain the addressee, to be funny. Metaphoric language is exploited for its divertive effects, which would fade in literal paraphrases.

\pex \label{ex: Humour}
\a I’m \underline{a doormat in the world of boots}.
\a You walked into what I would call \underline{a cupboard} but they classed it as the bathroom.
\xe 

Language is not only used to communicate. Among the many and varied uses of language, there is also the one of entertaining others, and being entertained in return. \citet{Steen2008, Steen2014} cites typical cases of humorous metaphors: sports newspaper headers, jokes, riddles and so on. In fact, the expression ”humorous metaphor” could stand for an umbrella concept grouping different phenomena, as suggested by \citet{attardo-2015-humorous}. The \textit{Resolvable Incongruity} view offers a possible explanation for the divertive potential of certain metaphors \cite{Oring2003, Dynel2009}.

\paragraph{Heuristic reasoning.} The intention is to provide an interpretative model for a theory, an artwork, etc., typically an abstract domain which is otherwise difficult to structure and conceive of. The metaphoric expression is used to organize the addressee’s conceptualization of the TARGET, based on their prior knowledge about the SOURCE. The discourse generally remains among specialists.

\pex \label{ex: Heuristic_reasoning}
\a A gas is \underline{like a collection of billiard balls} \underline{in random motion}.
\a It is her body \underline{as the canvas}, her appearance \underline{as art}.
\xe 

Metaphor is a matter of seeing something \textit{as} something else, that is, of interpreting things from a certain perspective. In cognitive terms, we map the SOURCE to the TARGET in order to better understand it. Thus, a primary intention of metaphor, especially within academic contexts, is to provide an interpretation for the products of science (\ref{ex: Heuristic_reasoning}a), as illustrated by Hesse in her seminal book \cite{Hesse1966}, or of art (\ref{ex: Heuristic_reasoning}b) and literature \cite{Ricœur1975}.

\section{Data collection and annotation}

\subsection{Collecting the data}

In order to empirically test the proposed taxonomy, we collected and annotated data ($\sim$ 1.2k metaphors) from the VU Amsterdam Metaphor Corpus (VUAMC; \citealp{steen-etal-2010-vuamc})\footnote{\url{http://www.vismet.org/metcor/about.html}}. This freely-accessible corpus was chosen since it contains fine-grained metaphoricity annotations at word level; it includes different genres; it contains metaphors in different grammatical constructions; and it has been extended in subsequent work with other relevant annotations, such as metaphor novelty scores \cite{DD2018}. Metaphor-related words (MRWs) in the VUAMC are identified following the MIPVU identification procedure \cite{MIPVU}. The core idea behind the procedure is the distinction between \textit{contextual} and \textit{basic} meaning of words. Text fragments are collected from the British National Corpus (BNC) Baby \cite{bncbaby-2005}, a 4-million-words corpus of English language covering 4 registers (Academic, News, Fiction, Conversation). The VUAMC encodes multiple information at word level, including information on metaphor type, distinguishing among \textit{direct} and \textit{indirect} metaphors.

Direct metaphors are expressions whose dictionary
meaning coincides with the contextual meaning. For example, the word \textit{ferret} in the phrase \textit{he’s like a \underline{ferret}} is a direct metaphor.
Indirect metaphors, instead, are defined as expressions having a more basic dictionary meaning that differs from the contextual meaning. Consider, for instance, the use of \textit{valuable} in the sentence \textit{teachers do a \underline{valuable} work}.

Our corpus consists of 1214 MRWs collected from the VUAMC. We annotated all unique instances of direct metaphors found in the corpus (301 MRWs) and a subset of indirect metaphors (913 MRWs). The VUAMC contains redundant instances of the same direct metaphor--several MRWs correspond, \textit{e.g.}, to the phrase \textit{like a \underline{pi\~{n}ata} above the teeming streets of the city}. However, for the purpose of annotating intentions the most natural unit of analysis is the phrase since the same intention is typically attributed to all MRWs in it. Thus, for each direct metaphor we assigned an intention only to one MRW. Annotators manually selected which word to annotate, based on their intuition of which lexical unit contributes the most to the metaphoricity of the phrase.

To select a subset of indirect metaphors to annotate we used \citet{DD2018}'s novelty scores. We divided all indirect metaphors into 5 bins according to their novelty scores. We opted to focus only on the top two bins---MRWs with novelty scores in [1,0.6] or (0.6,0.2]---which correspond to the most novel metaphors. Our rationale was that more creative uses of metaphor would yield more interesting material for investigating intentions. Within these indirect metaphors, we annotated 913 MRWs.

Some further cases were excluded from the annotation of intentions:

\begin{itemize}
    \item Cases where there was \textit{not sufficient context} to fully interpret the metaphor and assign an intention. A wider context could in general facilitate annotation since the attribution of intentions is likely informed by the surrounding discourse. Example: ”\underline{contraption}!”
    \item Cases of \textit{idiomatic} use. Idiom is a kind of figurative language use that should be distinguished from metaphor. While idioms represent relatively fixed and stable expressions within a linguistic community, metaphors are more productive and can show variation. Example: ”Even so, no room to \underline{swing} a cat.”
    \item Some highly conventionalized \textit{interjections} were also excluded since, just like idioms, they do not seem to require any active metaphorical interpretation in terms of meaning transfer. Example: ”\underline{Bloody} hell!”
\end{itemize}  

Instances marked as cases to be excluded were not considered in subsequent study phases. The final dataset comprises 988 MRWs, each annotated with at least one intention from the taxonomy.

\subsection{Annotation procedure and guidelines}

The procedure for the annotation of direct and indirect metaphors consists of two key steps:

\begin{enumerate}
    \item The annotator should distinguish lexicalized metaphors from other types of metaphors. If they perceive some intention behind the metaphor other than pure communication of information to the receiver, then they shall move on to step 2.
    \item The annotator is asked to assign up to three intentions to the metaphor under analysis. In order to complete the task, they are provided with a table listing the taxonomic categories, each with its description and some  examples.
\end{enumerate}

The full guidelines can be found in Appendix \ref{sec:appendix}. In the guidelines, we provide a detailed description of the sequential steps to be followed during annotation. We also work out at length an example of annotation performed following the guidelines.

\subsection{Corpus annotation}

The annotation was carried out by an author of this paper, who was a Master's student in logic and philosophy of language.
In addition to the 9 intention categories in the taxonomy, we also include a ``dummy category'' to keep track of cases where an intention could not be attributed.

\paragraph{Inter-annotator reliability.}
Another author, a metaphor researcher, annotated a subset of the data (360 MRWs). This subset is representative of the whole annotated corpus and replicates its proportions between different metaphor types: direct metaphors, indirect metaphors with novelty score in [1-0.6] and in (0.6-0.2].

We calculate inter-annotator reliability for 301 of the 360 items, to which both annotators assign at least one intention category. Their agreement in terms of Krippendorff's $\alpha$ \cite{Poesio2008} is 0.77, indicating moderate-to-fair agreement. More details about the metric used can be found in Appendix \ref{appendix:agreement}.

\section{Corpus analysis}

We analysed our corpus to shed some light on the relationship between intentions and metaphor type, genre and novelty. Only the first attributed intention was considered for data analysis since no other intention was selected in most cases (827/988). Distribution of intention categories in the whole corpus and per metaphor type is shown in Table~\ref{type}, and further analysis of metaphor type can be found in Appendix \ref{appendix:type}. 

\begin{table}[!t]
\begin{adjustbox}{width=0.48\textwidth}
\begin{tabular}{lcccc}
    \hline
    & Direct & Indirect [1,.6] & Indirect (.6,.2] & \textbf{Total} \\
    \hline
    Lexicalized metaphor & 9 & \fcolorbox{red}{white}{19} & \fcolorbox{red}{white}{379} & \fcolorbox{red}{white}{407} \\
    Artistic metaphor & \fcolorbox{orange}{white}{19} & \fcolorbox{yellow}{white}{13} & 43 & \fcolorbox{yellow}{white}{75} \\
    Visualization & \fcolorbox{red}{white}{53} & \fcolorbox{yellow}{white}{11} & \fcolorbox{orange}{white}{132} & \fcolorbox{orange}{white}{196} \\
    Persuasiveness & 2 & \fcolorbox{orange}{white}{15} & \fcolorbox{yellow}{white}{51} & \fcolorbox{yellow}{white}{68} \\
    Explanation & 9 & 3 & 30 & 42 \\
    Argumentative metaphor & 4 & 7 & \fcolorbox{yellow}{white}{48} & 59 \\
    Social interaction & 5 & 2 & 26 & 33 \\
    Humour & \fcolorbox{yellow}{white}{12} & 10 & 28 & 50 \\
    Heuristic reasoning & \fcolorbox{yellow}{white}{16} & 3 & 39 & 58 \\
     \hline
\end{tabular}
\end{adjustbox}
\caption{Distribution of intentions by metaphor type.}
\label{type}
\end{table}

\begin{figure}[!t]  \centering  \includegraphics[scale=0.4]{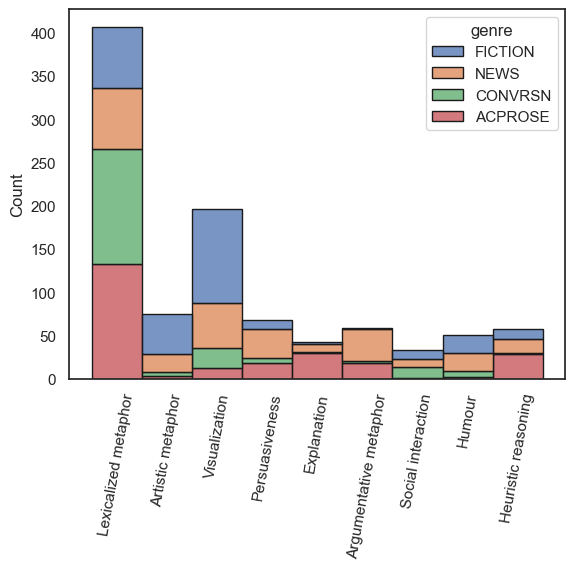}
    \caption{Distribution of intention categories per genre.}
    \label{genre}
\end{figure}

\begin{figure*}[!t]
    \includegraphics[scale=0.375]{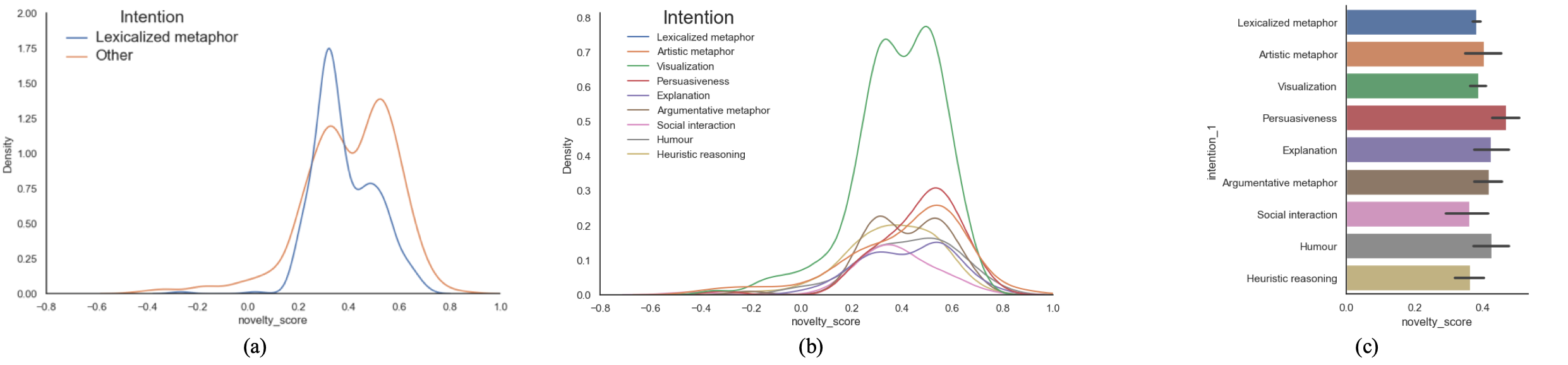}
    \caption{Distribution of novelty scores: (a) comparison between Lexicalized \textit{vs} other metaphors, (b) individual distributions, and (c) mean novelty scores. Figures (a), (b) show probability densities.}
    \label{novelty}
\end{figure*}

\paragraph{Genre.}

The genre of the discourse in which a metaphor appears should intuitively tell us something about its presumed intention. In particular, one would expect to find relatively more metaphors with a specific intention in extracts from certain genres, and not from others. This is suggested also by \citet{Steen2008}, who claims in passing that the function of a deliberate metaphor depends on the function of the discourse in which it is found. In the VUAMC, the information on the genre of each fragment is directly available under four intuitive tags: FICTION, NEWS, CONVRSN, ACPROSE.

In Figure \ref{genre}, we report how individual intention categories (the vertical bars) are distributed over the four genres (the coloured parts in each bar). Our findings support the assertion that intentions behind metaphor use seem to correlate with the discourse genre in which the metaphor is found. For instance, Artistic metaphor and Visualization are found mostly in Fiction; Persuasiveness and Argumentative metaphor in News; Explanation and Heuristic reasoning in Academic texts; Social interaction in Conversation. All of these results agree with what one would intuitively expect. However, reality is complex and suggests that drawing one-to-one correspondences would be too simplistic. In most cases, instances of the same intention are found in all four registers. Genre can thus help to track most common uses but not all uses.

\paragraph{Novelty score.} Information on the novelty \textit{vs} conventionality of metaphors is crucial for understanding how different intentions are reflected in different language choices. In particular, certain intentions seem to correlate with highly conventional metaphors, while others result in more original ones. 

Figure \ref{novelty}(a) contrasts the distribution over Lexicalized metaphor (the blue line) \textit{vs} all other intentions merged together (the orange line). In Figure \ref{novelty}(b), we zoom in and plot individual distributions. Each coloured line corresponds to the distribution of a single intention category.
Finally, we have computed mean novelty scores per intention with confidence intervals, as shown in Figure \ref{novelty}(c).

In terms of novelty, metaphors with different perceived intentions show different degrees of conventionality. Taking into account average novelty scores and estimated distributions, categories such as Persuasiveness, Explanation, Humour and Artistic metaphor are generally more original, while Lexicalized metaphor, Social interaction and Heuristic reasoning are more conventional.

\section{Evaluation of LLMs}

We use our dataset to test GPT-4 Turbo (\texttt{gpt-4-0125-preview};  \citealp{openai-2023-gpt4}), Llama2-13B-Chat, and Llama2-70B-Chat \cite{touvron-etal-2023-llama}
in terms of their ability to predict the intentions behind metaphor use (for details regarding model access, parameters, and computational budget, see Appendix~\ref{appendix:model}).
The task requires the models to choose a single intention category from our taxonomy, given a highlighted metaphorical expression in a sentence.
We test the models in zero-shot and five-shot in-context learning settings.
In the zero-shot setting, a short explanation for each intention category is provided.

In the five-shot settings, we randomly sample five in-context examples for each test item, and at least one of the examples is from the same intention category as the test item.
Since the in-context examples implicitly explain the intention categories,
we conduct two five-shot experiments:
One experiment provides the same explanations for the intention categories
that are used in zero-shot experiments (5-shot);
the other removes those explanations from the prompt (5-shot-short).
The latter setup tests whether the models are able to correctly infer what each intention category means from in-context examples. 
For each setting, we compute the average performance of the models across 3 different prompts, as shown in Appendix~\ref{appendix:prompts}.

\paragraph{Results.}
Table~\ref{tbl:acc} shows the models' performance in these tasks in terms of accuracy. 
All three models reach accuracies that are above the random baseline in the zero-shot experiments, although the accuracies are still relatively low, demonstrating that this is a challenging task for the LLMs.
GPT-4 is the best-performing model among the three,
and the 70B Llama2-Chat model slightly outperforms the 13B one.

In the 5-shot experiments, the accuracy of GPT-4 and Llama2-13B-Chat increases when the explanations for intention categories are retained in the instructions.
When the explanations are removed, the performance of all three models decreases, indicating that the models are not able to infer a correct characterization of the intention from the examples.
The accuracy of Llama2-70B-Chat is very close to the random baseline
in the 5-shot-short setting,
indicating that the model is particularly dependent on the explanations
in making correct predictions. 

\begin{table}[!t]
\centering
\small
\begin{tabular}{lrrr}
\hline
           & 0-shot       & 5-shot & 5-shot-short \\
\hline
Llama2 & & & \\
\texttt{-13b-chat} & 24.79 (2.42) & 26.75  & 22.90        \\
\texttt{-70b-chat} & 27.29 (5.45) & 22.49  & 14.39        \\
GPT-4 & 42.99 (1.64) & 44.68  & 41.44        \\
\hline
Random     & & & 13.01 \\
\hline
\end{tabular}
\caption{\label{tbl:acc} Model accuracy (\%) in zero- and few-shot settings, compared to random baseline.
Zero-shot accuracy is averaged over 3 runs that use different prompts;
standard deviation is given in parentheses.
The 5-shot-short setting removes explanations for intention categories from the prompt.}
\end{table}

\begin{figure}[!t]
    \includegraphics[scale=0.35]{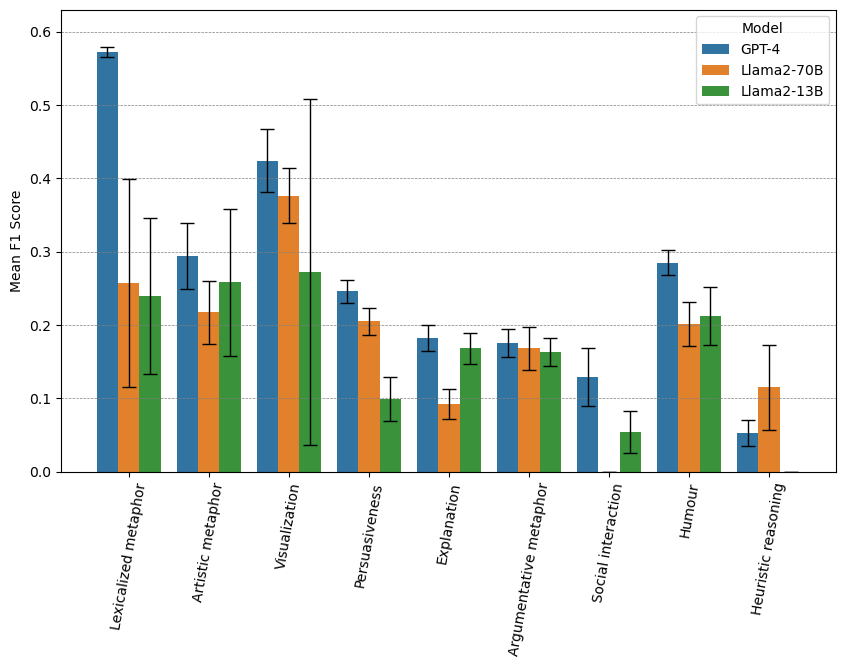}
    \caption{Model F$_1$ score in the zero-shot experiment, averaged across three prompts. Confidence intervals are computed with standard deviation across the prompts.}
    \label{0_shot}
\end{figure}

\paragraph{Error analysis.}
Figure \ref{0_shot} shows the mean F$_1$ score for each intention category in the zero-shot experiment, averaged across the three prompts.
GPT-4 reaches the highest F$_1$ scores when it comes to Lexicalized metaphor and Visualization, closely followed by Llama2-70B-Chat with regard to Visualization.
On the other hand, GPT-4 unsurprisingly gets the worst results in the Heuristic reasoning and Social interaction categories--recall from Section 5 that these are the least represented categories in our dataset.
F$_1$ scores for the 5-shot experiments are provided in Appendix~\ref{appendix:llm-results}.

Both GPT-4 and Llama2-70B-Chat mistake Lexicalized metaphor as Visualization.
These concerns conventional metaphors whose TARGET domains pertain to visible objects or the action of seeing, e.g., \lingform{a \underline{glimpse} of the impact of the 1980–1 riots}, and \lingform{\underline{channels} of communication}.

These two models also mistake Visualization for other intention categories, such as Artistic metaphor or Persuasiveness, e.g., \lingform{as enjoyable as \underline{feeling gently hungry or amorous}}; \lingform{the wide sleeves of limp cotton hung from her freckled arms like \underline{rags thrown over a stick}}.
These errors can be attributed to the absence of embodied experience in LLMs.
These metaphors naturally evoke a mental image or sensory experience in humans.
LLMs, on the other hand, do not automatically form representations of the meaning of a text in another modality.

\section{Conclusion}

\paragraph{Contributions.}

In our analysis, we have gathered evidence from the existing literature and incorporated it into a novel taxonomy of intentions commonly attributed to metaphor. The taxonomy can be used to annotate metaphors in unrestricted text, as demonstrated by our corpus annotation effort. Data collected from the VUAMC helped to better understand the nature of the different intentions and how these are realized in linguistic metaphors varying in their type, genre, and novelty score. Our taxonomy should be considered as a first step towards the systematization of the various findings on discourse goals that metaphor can accomplish. Individual categories can now be investigated further, and perhaps refined in future studies.

We have also created and released a first dataset with metaphors annotated according to the taxonomy.
Lastly, our experiments with GPT-4 and Llama2-Chat models show that inferring intentions behind metaphor use is still a challenging task for current LLMs.

\paragraph{Future directions.}

Since the VUAMC contains word-based annotations, we have currently adopted MRW as the basic unit for the annotation of intentions. A more natural choice, however, would be to analyse metaphoric phrases. Alternatively, the process of choosing a single MRW for each metaphoric expression could be automatized, for instance exploiting semantic information available at word level such as novelty, imageability or concreteness scores \cite{Wilson1997}. We intend to explore both routes in future work.

\section*{Limitations}

This study inevitably has some limitations; we discuss three of them here. First, the corpus used for the annotation, the VUAMC, contains mostly indirect metaphors, which are generally quite conventional. Adopting a corpus with more direct and novel metaphors would probably yield interesting results in terms of attributed intentions. However, such a corpus, comparable in size and range to the VUAMC, is missing. Second, while the output of the reliability study is encouraging, there is the need for extra annotators to make our results more trustworthy. Third, the current experimental setup asks LLMs to select only one intention category per metaphor, in contrast with our annotation guidelines. While making the task more straightforward, this choice does not reflect the complexity inherit to the analysis of metaphors in language use.

\section*{Ethical considerations}

Our dataset is created from metaphors sampled from the VUAMC, which is freely and publicly accessible and suitable for research purposes.
The two annotators are authors of this paper and volunteered to annotate the dataset.

\bibliography{anthology,refs_gmichelli,refs_xtong}
\bibliographystyle{acl_natbib}

\appendix

\section{The annotation guidelines}
\label{sec:appendix}

In this task, you are asked to annotate the intentions behind direct and indirect metaphors. For each sentence you are presented with, please annotate the text delimited by $<$b$>$ and $</$b$>$. For instance, in the sentence "Usually the slightest whisper travelled like jungle $<$b$>$drums$</$b$>$ through the world of fashion" you should annotate the word "drums", following the steps that are detailed below.

\begin{itemize}
\item \textbf{Step 1}: \underline{decide if the metaphoric expression} \underline{could be avoided}.\\
If there are (literal) paraphrases that would convey roughly the same message in the given context, please continue the annotation and proceed with Step 2. If you cannot think of any paraphrase that avoids the metaphor and would work just fine, then mark the metaphor as \textit{Lexicalized metaphor} and skip Step 2.

\item \textbf{Step 2}: \underline{select categories from the taxonomy} \underline{of intentions}.\\
In this step, you are asked to select a possible intention behind the metaphor you are analysing. The list of categories that you should use is the following one: \textit{Artistic metaphor}, \textit{Visualization}, \textit{Persuasiveness}, \textit{Explanation}, \textit{Argumentative metaphor}, \textit{Social interaction}, \textit{Humour}, \textit{Heuristic reasoning}. If you think that more intentions might play a role, feel free to select multiple categories--up to a maximum of 3. 
\end{itemize}

\subsection{Explanation}

\textbf{Lexicalized metaphors.} To discriminate between lexicalized metaphors and other metaphors, try to think about the subject matter (the Topic) of the metaphor. If the metaphor is just the most common way to talk about the Topic, then mark it as \textit{Lexicalized metaphor}. On the other hand, if the metaphor could be avoided, and the intended message could be expressed in a different way, then the metaphor is not lexicalized. Consider the following examples:

\pex \label{ex: Step1}
\a Do you $<$b$>$follow$</$b$>$?
\a Usually the slightest whisper travelled like jungle $<$b$>$drums$</$b$>$ through the world of fashion.
\xe

(\ref{ex: Step1}a) is an example of a lexicalized metaphor. The speaker is asking the hearer if they are "following" (most likely) their words. This simply reflects the way in which we generally conceptualize discourse, namely in spatial terms (\textit{e.g.} as a path).

On the other hand, the metaphor in (\ref{ex: Step1}b) is not lexicalized. The noun "drum" is not commonly used to talk about fashion. One could express the intended message through the following paraphrase "Usually the slightest whisper spread very fast and loud though the world of fashion".\\

\textbf{Intention categories.} For Step 2, try to think of which communicative goals the metaphor might accomplish better than its paraphrases. To decide which intention(s) to select, refer to the following overview of the taxonomic categories. Each item is provided with its description and some paradigmatic examples.\\

\newgeometry{}

\small
\noindent \begin{tabular}{ |V{4cm}|V{4cm}|V{6.5cm}| }
\hline
\textbf{Intention} & \textbf{Description} & \textbf{Examples} \\
\hline
\vspace{0.05cm} 
\vspace{0.05cm} \textit{Artistic metaphor} & \vspace{0.05cm} \small These metaphors are used to predicate at once a whole set of features of the Topic. These features need not to be all clearly determined in advance. Ultimately, the intention is to stimulate the receiver's creative interpretation. & \small \begin{itemize} \setlength\itemsep{0.5mm}
\item  To her, the long summer days had stretched ahead, $<$b$>$world$</$b$>$ without end. 
\item Amaldi dodged the American invitation, perhaps because (with Rome liberated) Fermi's $<$b$>$mantle$</$b$>$ in physics had fallen on his young shoulders and there were younger minds to teach.
\item The summer's $<$b$>$sprawl$</$b$>$ begins to be oppressive at this stage in the year and trigger fingers are itching to snip back overgrown mallows, clear out the mildewing foliage of golden rod and reduce the overpowering bulk of bullyboy ground cover.
\end{itemize}
\\
\hline

\vspace{0.05cm} \textit{Visualization} & \vspace{0.05cm} \small The utterer might resort to a metaphor whose Vehicle (\textit{i.e.} the conventional referent) is easier to visualize than the Topic (the contextual referent). Typically, this happens when the latter belongs to an abstract domain or when the audience is not familiar with it. The intention is to help the receiver to form an intuitive representation of the Topic. & \small \begin{itemize} \setlength\itemsep{0.5mm}
\item Relief surged through her like a physical $<$b$>$infusion$</$b$>$ of new blood.
\item And beyond, green grass and geraniums like $<$b$>$splashes$</$b$>$ of blood.
\item The results are terse and sharply $<$b$>$etched$</$b$>$, like the best line drawings.
\end{itemize}
\\
\hline

\textit{Persuasiveness} &  \vspace{0.05cm} \small Using the metaphor to refer to the Topic, the author gives it a non-neutral connotation, which is not motivated on explicit grounds. The intention is for the audience to adopt the utterer's positive or negative attitude towards the Topic. & \small \begin{itemize} \setlength\itemsep{0.5mm}
\item The $<$b$>$ramshackle$</$b$>$ Whitley Council negotiating machinery is the other reason why the ambulance workers have lost out. 
\item America may have changed Presidents a year ago, but the fiscal ticket remains as $<$b$>$inpenetrable$</$b$>$ as ever.
\item An atmosphere $<$b$>$poisoned$</$b$>$ by mistrust. \end{itemize}
\\
\hline

\vspace{0.05cm} \textit{Explanation} & \vspace{0.05cm} \small These metaphors are used for didactic purposes. The intention is to explain a new or already familiar concept to the addressee. & \small \begin{itemize} \setlength\itemsep{0.5mm}
\item Canals within the algae stand out as $<$b$>$rods$</$b$>$ in this kind of preservation, which is common in Ordovician rocks.
\item Thus one can and must say, that each fight is the singularisation of all the circumstances of the social whole in movement and that by this singularisation, it $<$b$>$incarnates$</$b$>$ the enveloping totalization which the historical process is.
\item The ego-identity of that person is $<$b$>$shaped$</$b$>$ by these choices.
\end{itemize}
\\
\hline

\end{tabular}

\begin{tabular}{ |V{4cm}|V{4cm}|V{6.5cm}| }
\hline

\vspace{0.05cm} \textit{Argumentative metaphor} & \vspace{0.05cm} \small These metaphors are part of explicit arguments intended by the author to convince the audience of a certain claim. The intention is to support the argument, to make it more compelling for the addressee. & \small \begin{itemize} \setlength\itemsep{0.5mm}
\item The effect is rather like an extended $<$b$>$advertisement$</$b$>$ for Marlboro Lights.
\item There was already a rather perfunctory air to the Queen's visit three years ago, as if it were just a required $<$b$>$coda$</$b$>$ to her tour of China.
\item But the villages are dying, becoming suburbs or $<$b$>$dormitories$</$b$>$ where few people work but many sleep.
\end{itemize}
\\
\hline

\vspace{0.05cm} \textit{Social interaction} & \vspace{0.05cm} \small These metaphors focus on interpersonal relations, group or cultural conventions and the like. The intention is to create or strengthen some bond between producer and receiver. & \small \begin{itemize} \setlength\itemsep{0.5mm}
\item But I'm starting to think that everything's a turn-off for you, $<$b$>$doll$</$b$>$.
\item Smoking heroin ("$<$b$>$chasing$</$b$>$ the dragon") was one feature of the upsurge.
\item Political correctness, just as we suspected, will be perfectly $<$b$>$grey$</$b$>$.
\end{itemize}
\\
\hline

\vspace{0.05cm} \textit{Humour} & \vspace{0.05cm} \small The intention is to entertain the addressee, to be funny. Metaphoric language is exploited for its divertive effects, which would go missing in literal paraphrases. & \small \begin{itemize} \setlength\itemsep{0.5mm}
\item  Not sure of the music policy, but the name sounds like the $<$b$>$ingredients$</$b$>$ of a takeaway from a less salubrious Chinese. 
\item From there, like a $<$b$>$buzzard$</$b$>$ in its eyrie, he would make forays round the US and abroad in spite of his advanced age.
\item It 's my life which is about to go down the $<$b$>$plughole$</$b$>$.
\end{itemize}
\\
\hline

\vspace{0.05cm} \textit{Heuristic reasoning} & \vspace{0.05cm} \small The intention is to provide an interpretative model for a scientific theory, a work of art, etc. The metaphoric expression is used to organize the addressee's conceptualization of the Topic, based on their prior knowledge about another domain.  & \small \begin{itemize} \setlength\itemsep{0.5mm}
\item  It is her body as the $<$b$>$canvas$</$b$>$ her appearance as art.
\item It is as if it is walking through a $<$b$>$minefield$</$b$>$.
\item At the moment, history is made without being known (l'histoire se fait sans se connaître); history constitutes, we might say today, a political $<$b$>$unconscious$</$b$>$.
\end{itemize}
 \\
\hline

\end{tabular}

\normalsize
\vspace{0.2cm}
\restoregeometry
\newpage
\subsection{Example}

Here below is one example annotated following the guidelines.\\

Allan Ahlberg says: "In the past, a lot of children's books seemed to be the work of talented illustrators whose pictures looked brilliant framed in a gallery, but when you tried to read the book, there was nothing there, because the words started as a $<$b$>$coat-hanger$</$b$>$ to hang pictures on."\footnote{The example is taken from a News text in the VUAMC (document id: a1l-fragment01; sentence id: 29).}\\

\textbf{Step 1}. This sentence from a news fragment is about old children's books. The author highlights the characteristic of these books of focusing more on the quality of the illustrations, rather than on the narration. The words that make up the story are metaphorically compared to coat-hangers. The utterer invites us to think of the relation between the illustrations and the words as the one existing between a coat and a coat-hanger. The latter is just instrumental, it has no purpose or value in itself which is independent of the former. Through the metaphor, the author predicates these features of the words in the children's books. The same message could have been conveyed in a literal way, along the following lines: "the words had no value in themselves, they were just instrumental for the illustrations". Thus, the output of Step 1 is that the metaphor is \textit{not} \textit{lexicalized} and we may move on to Step 2.\\

\textbf{Step 2}. The metaphoric expression is used in this case to explain the way in which illustrations and words are related in old children's books. The author invites the addressees to understand this relation in terms of the more familiar and concrete relation between coats and hangers. For this reason, the metaphor can be annotated as \textit{Explanation}. It should be noted, however, that also other intentions seem to play a role. For instance, one might read a negative judgment of value in the author's remark. Thus, the annotation could also be \textit{Persuasiveness} or \textit{Argumentative metaphor}, depending on whether some rational justification is given by the utterer to support their judgment.

\section{Inter-annotator agreement}
\label{appendix:agreement}

Our annotation task consists of a multi-label classification with multiple annotators--individual instances can be associated with multiple, non-exclusive intentions. After a brief survey of the available options \cite{Poesio2008}, we opted for a variant of Krippendorff's $\alpha$ as an indicator of the inter-annotator agreement. In particular, we adopted the MASI distance, which is suitable for set-valued labelling tasks such as ours\footnote{The metric has been applied by Passonneau and colleagues to the annotation of co-reference chains \cite{Passonneau2004} and Summary Content Units \cite{Passonneau2006}.}.

Out of the 360 MRWs included in the reliability study, 59 distinct items were judged as cases to be excluded by either or both of the two coders. Inter-annotator agreement was computed on the remaining 301 metaphors, where at least one intention was assigned by each annotator. The inter-annotator agreement score was 0.77, which indicates moderate-to-fair agreement. While in his seminal work \citet{Krippendorff1980} sets 0.8 as the minimal requirement for reliable annotation schemes, we agree with \citet{Poesio2008}, who "doubt that a single cutoff point is appropriate for all purposes" and indicate 0.7 as a more reasonable goal, especially for complex semantic tasks.

\section{Corpus analysis: Type}
\label{appendix:type}

Proponents of DMT maintain that direct metaphors constitute principled examples of deliberate metaphors. Since direct metaphors overtly introduce a referent from a SOURCE domain from which a conceptual mapping has to be made \cite{Steen2011}, they would require the intentional use of metaphor \textit{as} metaphor. On the contrary, given the availability of a contextually relevant non-basic meaning, indirect metaphors would be non-deliberate--though ambiguous cases are possible \cite{Steen2023}. Thus, information on the type of linguistic metaphor would help to identify deliberate uses in communication. In Table \ref{type}, we outline the distribution of metaphors in our dataset across the intention categories for all metaphor types.

The results partially align with the claim that direct and indirect metaphors show different tendencies when it comes to their perceived intentions. While all meaningful metaphors are uttered with the minimal intention to communicate, direct metaphors generally correlate with other discourse goals, too. The categories mostly associated to direct metaphors are Visualization, Artistic metaphor, Heuristic reasoning. Indirect metaphors, especially the most conventional ones, are instead judged as lexicalized metaphors.

\section{Model details}
\label{appendix:model}

The GPT-4 model is accessed through the OpenAI API,
and the two Llama2-Chat models Hugging Face.
We employ greedy search for all 3 models.
For the two Llama2-Chat models, this is done by setting
\texttt{do\_sample=False} and \texttt{num\_beams=1};
for the GPT-4 model, \texttt{temperature} is set to 0.

Our GPT-4 queries cost $\sim 60$ USD.
Our Llama2-Chat queries used $\sim 460$ GPU hours (58946:35 SBU).

\section{Prompts}
\label{appendix:prompts}

\begin{figure*}
\includegraphics[width=\textwidth]{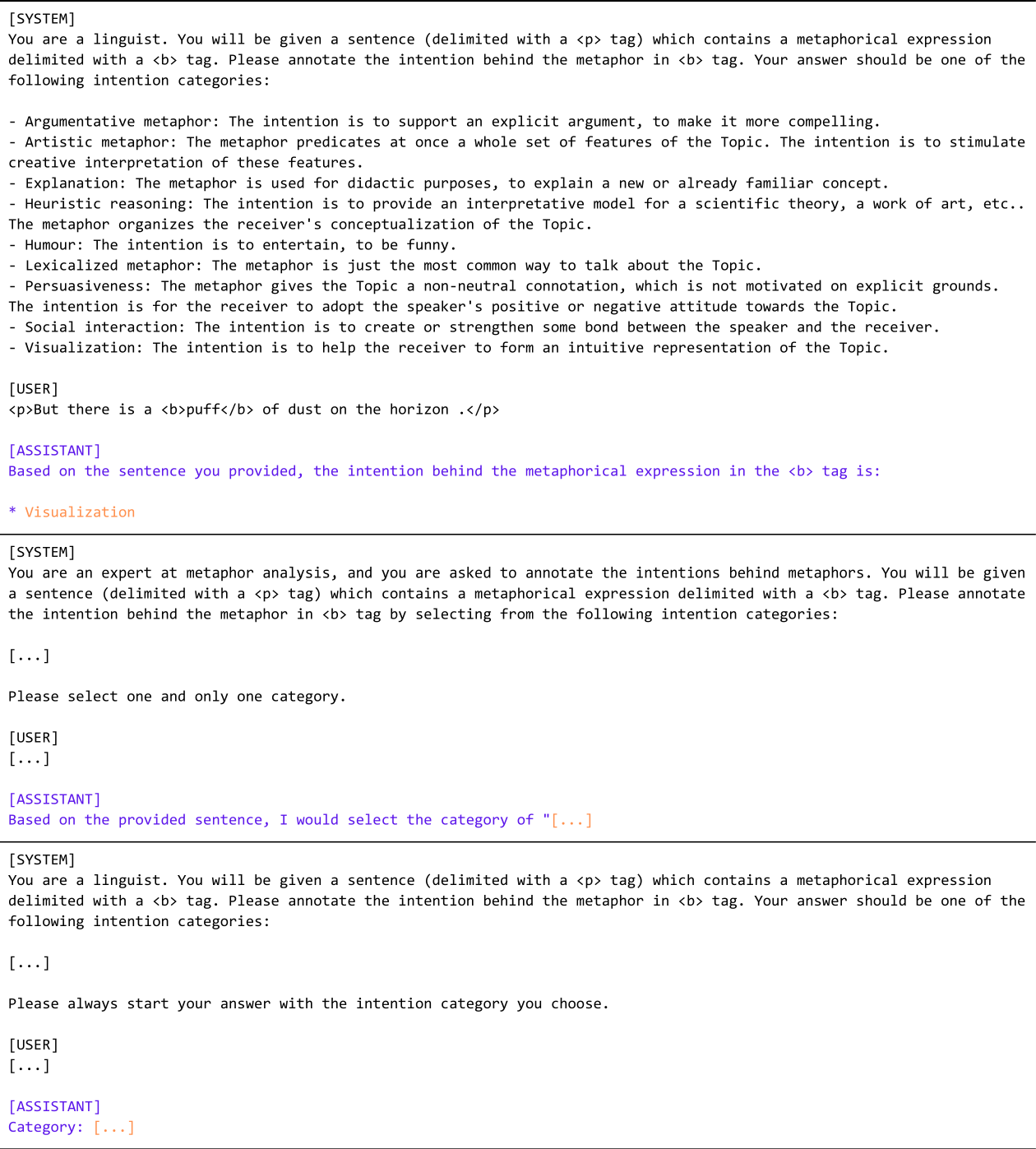}
\caption{\label{fig:prompts-0shot}
Prompts for zero-shot experiments and \textcolor{promptOrange}{example model output}. The same explanations for the intention categories are used in all 3 prompts. Assistant messages in violet are provided to the \textcolor{promptViolet}{Llama2-Chat} models, so that model outputs always start with the predicted intention category.} 
\end{figure*}

\begin{figure*}
\includegraphics[width=\textwidth]{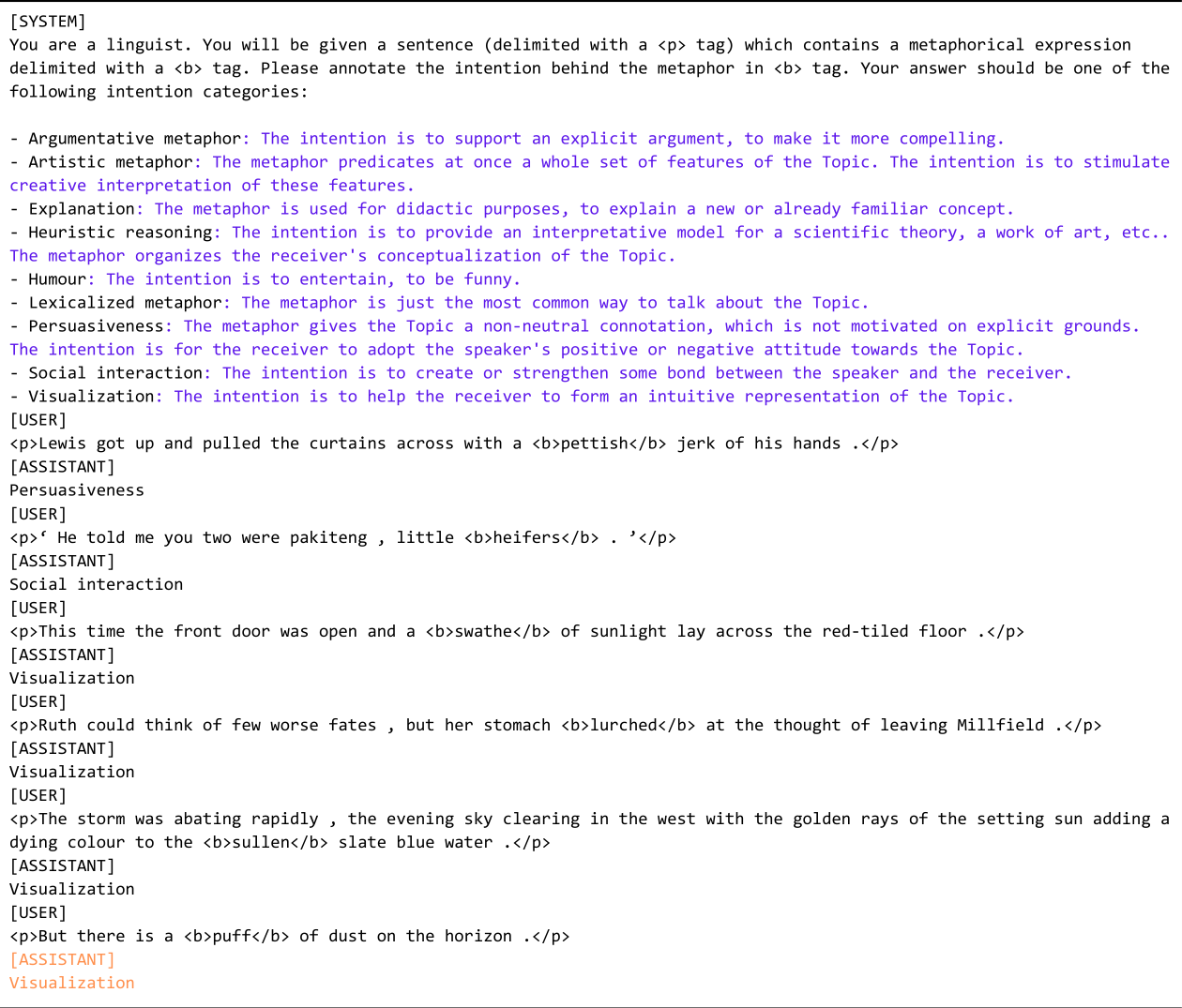}
\caption{\label{fig:prompts-5shot}
Prompts for five-shot experiments and \textcolor{promptOrange}{example model output}. \textcolor{promptViolet}{The explanations for intention categories are removed in the 5-shot-short setting.}} 
\end{figure*}

The prompts for zero-shot and five-shot experiments are presented in Figure~\ref{fig:prompts-0shot} and \ref{fig:prompts-5shot} respectively.
In the zero-shot experiments, the GPT-4 model always starts its answer with the intention category it predicts for the given metaphor.
The Llama2-Chat models, on the other hand, need to generate some text (for example, \lingform{Based on the provided sentence, I would select the category of \ldots}) before providing its prediction.
We thus provide the Llama2-Chat models the text they tend to generate at the start of their assistant messages (as part of the prompts), so that the first few new tokens they generate will be the intention category they predict.

Such assistant prompts are determined in the following way:
We first take a prompt (system message and user message) that works for GPT-4 and apply it directly to a Llama2-Chat model (the 13B model for the first 2 prompts, and the 70B model for the last one).
We do this for 3 different input sentences to obtain the text the model is most likely to produce before providing its prediction.
This text is then used as the assistant prompt for both Llama2-Chat models.
As shown in Figure~\ref{fig:prompts-0shot}, the 3 prompts contain different assistant messages, as we follow the messages that the Llama2-Chat models naturally produce when provided with different system prompts.

\section{Model performance}
\label{appendix:llm-results}

\begin{figure}[!t]
    \includegraphics[scale=0.35]{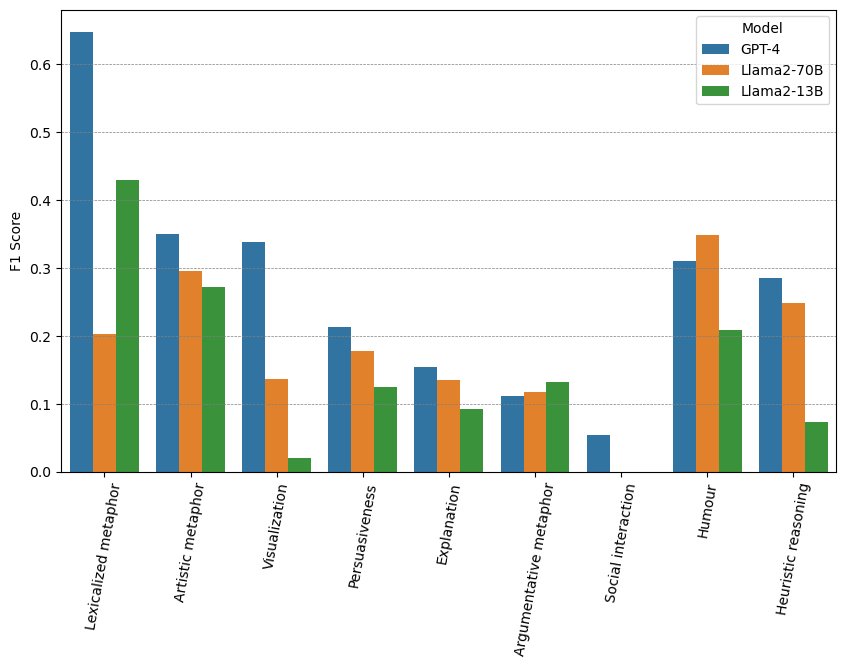}
    \caption{Model F$_1$ score in the five-shot experiment with explanations for the intention categories.}
    \label{5_shot}
\end{figure}

\begin{figure}[!t]
    \includegraphics[scale=0.35]{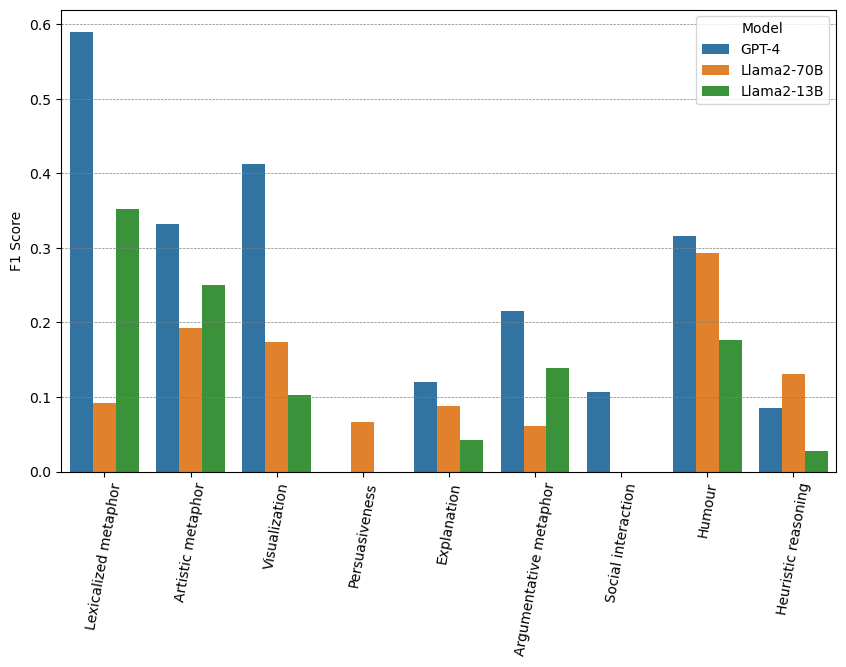}
    \caption{Model F$_1$ score in the five-shot experiment without explanations for the intention categories.}
    \label{5_shot_short}
\end{figure}

Figures~\ref{5_shot} and \ref{5_shot_short} show the three models' performance (F$_1$ scores) in the 5-shot settings with regard to each intention category.

\end{document}